\documentclass[11pt]{article}
\usepackage{coling2020}
\usepackage{times}
\usepackage{url}
\usepackage{latexsym}
\usepackage{graphicx}
\usepackage{multirow}
\usepackage{amsmath}
\usepackage{mathtools}
\usepackage{tabularx}
\usepackage{url}
\usepackage{subcaption}
\usepackage{tikz-qtree}
\usepackage{pgfplots}
\usepackage{latexsym}

\newcommand{\comment}[1]{}
\usepackage{microtype}

\colingfinalcopy 

\title{From Sentiment Annotations to Sentiment Prediction \\through Discourse Augmentation 
}

\author{Patrick Huber and Giuseppe Carenini\\
  Department of Computer Science \\
  University of British Columbia \\
  Vancouver, BC, Canada, V6T 1Z4 \\
  {\tt \{huberpat, carenini\}@cs.ubc.ca}}

\date{}

\begin{document}
\maketitle
\begin{abstract}
Sentiment analysis, especially for long documents, plausibly requires methods  
capturing 
complex linguistics 
structures. 
To accommodate this, we propose a novel framework to exploit task-related discourse for the task of sentiment analysis. More specifically, we are combining the large-scale, sentiment-dependent MEGA-DT treebank 
with a novel neural architecture for sentiment prediction, based on a hybrid TreeLSTM hierarchical attention 
model.
Experiments show that our 
framework using sentiment-related discourse augmentations for sentiment prediction enhances the overall performance for long documents, even beyond previous approaches using well-established discourse parsers trained on human annotated data. We show that a simple ensemble approach can further enhance performance by selectively using discourse, 
depending on the document length.

\end{abstract}

\section{Introduction}
\label{intro}
\blfootnote{
    \hspace{-0.65cm}  
    This work is licensed under a Creative Commons 
    Attribution 4.0 International License.
    License details:
    \url{http://creativecommons.org/licenses/by/4.0/}.
}
Predicting whether a given word, sentence or document expresses a positive, neutral or negative sentiment is a fundamental task in Natural Language Processing (NLP). For instance, a recent survey of text mining papers from 1992-2017 has found that out of $4,346$ papers, $467$ had a sentiment analysis component \cite{Bridging}. While early ``bag-of-word" sentiment prediction models \cite{taboada2011lexicon} and their extensions \cite{wilson-etal-2009-articles} already show promising results on the task, they all share one inherit limitation: Due to the absence of temporal information, they are not able to fully capture the semantics (and therefore the sentiment) of long texts, where different meanings oftentimes directly emerge from the word order, underlying syntax and discourse structures. 

Recent models for sentiment analysis address this limitation by leveraging sequential paradigms \cite{dos2014deep,kim2014convolutional,tai2015improved,adhikari2019rethinking}, simple hierarchical information \cite{yang2016hierarchical}, complex syntactic structures on sentence level \cite{socher2013recursive} or discourse structures of multi-sentential text \cite{ji2017neural}.

\begin{figure}[t!]
\centering
    \begin{subfigure}{.5\textwidth}
    \centering
    \includegraphics[width=.85\linewidth]{ 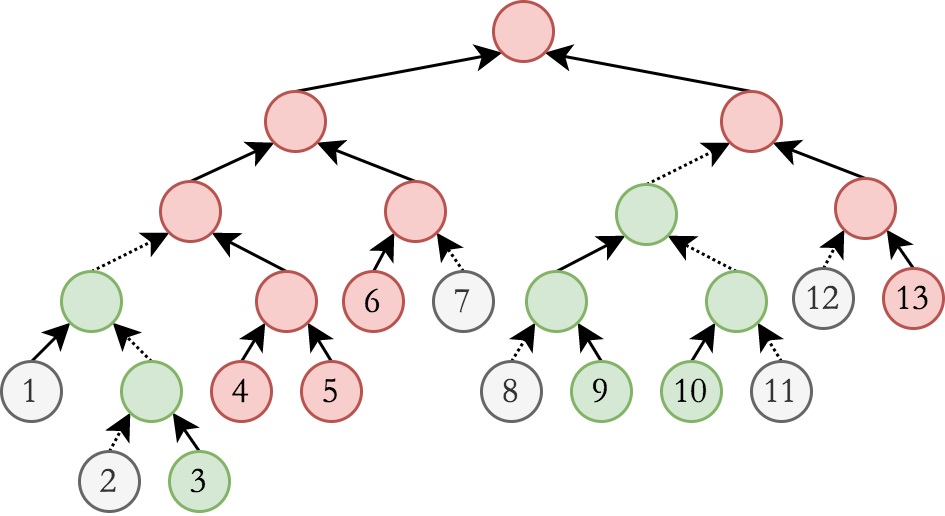}
    \end{subfigure}%
\begin{subfigure}{.5\textwidth}
  \centering
    \includegraphics[width=\linewidth]
    {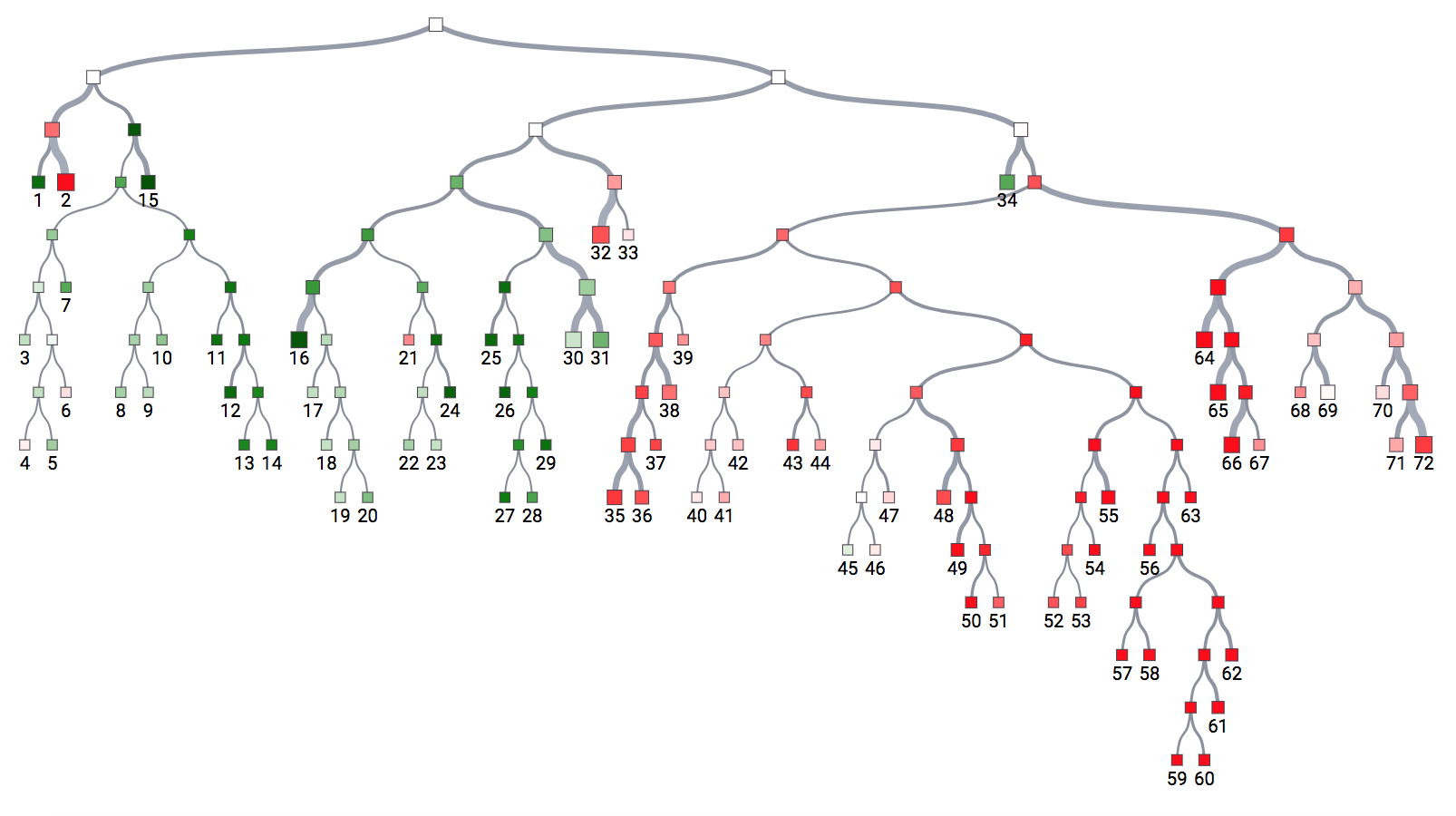}
\end{subfigure}
\caption{Sentiment annotated discourse trees for non-trivial documents containing 13 (left) and 72 (right) clause-like components with positive and negative constituents. Gold-label sentiment is negative (left) and neutral (right). Dashed/Thin lines indicate supplementary information, solid/thick lines indicate primary importance. Full text for left example is: \textit{[I've been a member for a month now,]\textsubscript{1}, 
[and I guess]\textsubscript{2},
[I 'm able to get my workout done.]\textsubscript{3},
[I do find myself annoyed]\textsubscript{4},
[how cramped it is at the weights.]\textsubscript{5},
[The equipment is older,]\textsubscript{6},
[but it suffices.]\textsubscript{7},
[I worked out at another studio on 3rd]\textsubscript{8},
[and it was amazing!]\textsubscript{9},
[It was so clean, nice, and new -]\textsubscript{10},
[TV 's on every cardio machine.]\textsubscript{11},
[When i came back to this location,]\textsubscript{12},
[I felt bad.]\textsubscript{13}}
Due to space limitations a larger version as well as the corresponding full text for the right example is shown in Appendix \ref{appendix_1}.
}
\label{fig:example}
\end{figure}

This paper follows the last line of aforementioned research, by developing a framework to exploit automatically generated, large-scale, domain-related discourse structures for sentiment prediction. Arguably, such framework can be especially beneficial for long documents that examine positive and negative aspects of a subject matter in complex rhetorical structures, like the ones shown in Figure \ref{fig:example}.

More specifically, in this work, we generate complete and hierarchical RST-style discourse trees \cite{mann1988rhetorical} with leaf nodes representing  clause-like document fragments, called elementary discourse units (EDUs) and 
internal  tree nodes labelled with a nuclearity assignment (Nucleus, Satellite), encoding the importance of a node in its local subtree
\footnote{Discourse relation are not considered in this work.}. To incorporate these RST-style discourse structures, we employ a hybrid approach inspired by \newcite{bowman2016fast} and \newcite{choi2018learning}, integrating a TreeLSTM \cite{tai2015improved} with the well-established Hierarchical Attention Network model (HAN) \cite{yang2016hierarchical}. From \newcite{ji2017neural}, we further adopt a non-competitive tree attention mechanism that is shown to be more appropriate in this context\footnote{We did not apply tree-transformers to the task, as in spite of recent proposals (e.g. \newcite{shiv2019novel}, \newcite{nguyen2020tree}), no standard method has been widely agreed upon yet and results are still rather preliminary.}.

Aiming to enhance the task of sentiment analysis by using discourse, it seems intuitive to employ domain-related discourse structures. Therefore, instead of using the standard RST-DT discourse treebank in the news domain \cite{carlson2002rst}, we decide to infer discourse structures automatically learned from sentiment annotations \cite{huber2019predicting} on our discourse-augmented Yelp’13 treebank called MEGA-DT \cite{huber2020MEGA}. This way, our framework goes from sentiment to sentiment, in the sense that the discourse structures used to improve the sentiment predictions are generated through distant supervision from sentiment itself. Our hypothesis is that a parser trained on a large ``silver-standard" discourse treebank  automatically generated from sentiment will generate more useful discourse trees for sentiment prediction than one trained on a small and generic treebank, even if such treebank is human-annotated for RST discourse structures.

In a series of experiments we show that while our novel approach to discourse-based sentiment prediction is statistically equivalent to the performance of sequential models, it does deliver substantial performance gains for long documents, where discourse plays a crucial role to reveal the sentiment of a complete document. Furthermore, our experiments indicate that the performance of discourse-based sentiment prediction is significantly improved when using 
discourse trees generated by distant supervision on sentiment, 
compared to the traditionally acquired RST-DT discourse corpus. 
Using an additional ensemble method, we can further improve the performance and, even if only by a small margin, significantly outperform individual models. 

\section{Related Work}
\label{rel}
This work is located at the intersection of recent approaches on discourse parsing and sentiment analysis and mostly influenced by four lines of research:

\textbf{(1) RST-style Discourse Parsing} 
is a valuable upstream task for many downstream models (e.g. \newcite{ji2017neural}, \newcite{gerani2014abstractive}). 
Different approaches either separate discourse parsing 
``vertically" into sub-tasks on sentence-level, paragraph-level and document-level \cite{joty2015codra,ji2014representation}, or 
``horizontally", separating the prediction of structure and nuclearity from the relation computation \cite{wang2017two}. Furthermore, approaches have been explored to aggregate documents bottom-up using CKY \cite{joty2015codra} or employing local shift-reduce strategies, predicting the tree-structure through a sequence of actions based on linguistic features  \cite{ji2014representation,subba2009effective,wang2017two} or dense representations \cite{yu2018transition}.
Empirically, \newcite{wang2017two} show that the combination of horizontal separation with a shift-reduce parsing framework achieves competitive 
performance, reaching state-of-the-art results on the structure-prediction task. 
In this work, we  demonstrate the potential of this discourse parser trained on a large-scale sentiment-dependent treebank (MEGA-DT) to generate discourse trees for sentiment prediction, enhancing the performance on long and diverse documents.

\textbf{(2) Neural Sentiment Analysis} is a common sub-task in many real world systems with 
\newcite{kim2014convolutional} being the first to show the effectiveness of convolutional neural networks for the task. \newcite{yang2016hierarchical} followed shortly after with their Hierachical Attention Network model (HAN), proposing one of the first hierarchical models for text classification. HAN separates the task at the sentence-level and builds a model comprising of two hierarchical components, each with an additional attention mechanism. Further successful approaches to predict sentiment have been explored recently by \newcite{adhikari2019docbert}, proposing a model based on BERT, and \newcite{adhikari2019rethinking}, applying a simple but more regularized BiLSTM to the task.
In this fast moving area, our goal is to investigate the influence of discourse information on the task of sentiment analysis. We therefore decide to build our framework on the HAN model \cite{yang2016hierarchical}, which is the most established, yet recent approach in the field, previously re-implemented and tested in many studies. We inject discourse information using TreeLSTMs \cite{tai2015improved}, which are also well-established compared to tree-transformers, for which architectural variants and results are still preliminary (e.g. \newcite{shiv2019novel}, \newcite{nguyen2020tree}).

\textbf{(3) Combining  Discourse Parsing and Sentiment Analysis} has been previously explored in multiple lines of work \cite{bhatia2015better,hogenboom2015using,nejat2017exploring,ji2017neural}.
Architecture-wise, the most closely related approach to our new model has been proposed by \newcite{ji2017neural}, 
where discourse trees generated by the DPLP parser \cite{ji2014representation} trained on RST-DT are used in a recursive neural network to predict sentiment for multiple corpora. In their evaluation, the authors show slight improvements compared to the sequential HAN model. These initial positive results are a key motivation for our work, in which we aim to further improve the performance, especially on long documents, by not only training the discourse parser on a larger and more appropriate treebank (i.e. MEGA-DT), but also by improving the sentiment prediction, replacing recursive neural networks with superior TreeLSTMs, tightly integrated with HAN.

\textbf{(4) (Discourse) Tree Learning}
tries to automatically infer discourse trees from large amounts of data. In popular approaches, trees are inferred directly 
while learning a neural model for a downstream task, such as text classification \cite{karimi2019learning} or extractive summarization \cite{liu2019single}. 
Along this line of research, we previously proposed a similar objective in \newcite{huber2019predicting}, automatically generating discourse trees from distant supervision of a downstream task (sentiment analysis). However, we employed a rather different approach. Instead of trying to induce discourse trees directly during training of a neural network, we propose a dedicated system, comprising of well-established methods, to directly generate discourse trees. With the resulting large-scale, sentiment influenced discourse treebank called MEGA-DT, we reported promising results on the task of discourse parsing itself in \newcite{huber2020MEGA}. Showing the potential of applying MEGA-DT 
to the task of sentiment prediction is a goal of this work.\\

\section{Sentiment to Sentiment Framework
}
\label{integration}

\begin{figure*}
    \centering
    \includegraphics[width=1\linewidth]{ 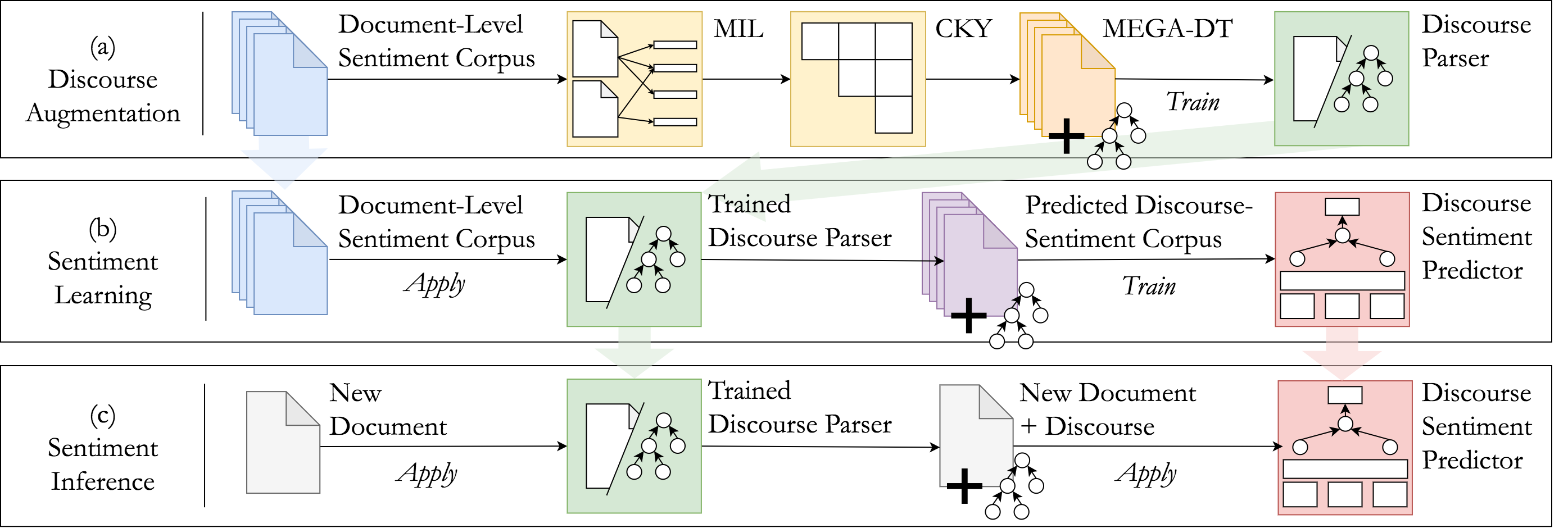}
    \caption{Our proposed sentiment analysis framework, containing, three phases of training/inference to integrate discourse parsing and sentiment analysis.}
    \label{fig:stages}
\end{figure*}

Our sentiment to sentiment framework involves three phases: A phase of discourse augmentation (Figure \ref{fig:stages} (a)), in which we follow our previous approach described in \newcite{huber2019predicting} and \newcite{huber2020MEGA}. For each document in a corpus containing document-level sentiment annotation, we generate corresponding, task-dependent discourse trees. Then, this discourse augmented sentiment treebank is used to train a discourse parser. In the second phase (Figure \ref{fig:stages} (b)),  the trained discourse parser is applied to the original corpus, using the predicted trees to train our new discourse-based sentiment predictor. Finally, in the third phase (Figure \ref{fig:stages} (c)), the trained framework is applied to any new document. First, the trained discourse parser generates the discourse tree for the document. Subsequently, this tree (along with the document itself) is fed to our sentiment predictor, which returns the most likely sentiment. In essence, we go from sentiment annotations to sentiment predictions through discourse augmentation.

For the first phase, we briefly describe the discourse augmentation step adopted from our previous work \cite{huber2019predicting,huber2020MEGA} in section \ref{sent_to_dis}. For phase two, we focus on our novel sentiment predictor in section \ref{dis_to_sent}. 
The inference phase is straightforward and will be limited to the description in Figure \ref{fig:stages} (c) for brevity.

\subsection{
Sentiment Inspired Discourse Trees}
\label{sent_to_dis}


The approach to generate ``silver-standard" 
partial discourse trees (incorporating structure and nuclearity) from distant sentiment supervision 
\cite{huber2019predicting,huber2020MEGA} comprises two major components. First, documents are annotated for sentiment and importance at the EDU-level using a neural Multiple-Instance Learning (MIL) method \cite{angelidis2018multiple}, solely utilizing document-level supervision signals given in the original corpus. In particular, MIL infers a sentiment polarity label $p_x$ within the interval of $[-1,1]$ for each EDU $x$, depending on the distribution of words/EDUs within and between documents. 
Using the neural model by \newcite{angelidis2018multiple}, an additional attention mechanism is internally used to weight the importance of EDUs for the overall document sentiment. The attention-weight $a_x$ in the interval $[0,1]$ of EDU $x$ is also extracted from the model and subsequently used as an importance score when aggregating sub-trees.
Next, the tuples $(p_x, a_x)$ 
are combined in a binary, bottom-up approach using dynamic programming, inspired by CKY \cite{jurafsky2014speech}. 
With a multitude of possible discourse trees generated in this way, the tree-structure minimizing the divergence between the document sentiment gold-label and the predicted sentiment, obtained by combining the tuples $(p_x, a_x)$ according to equation \ref{eq:aggregation}, is deemed to represent the document discourse-structure.

\vspace{.1cm}
\begin{equation}
p = \frac{ p_{c_{l}}*a_{c_{l}}+p_{c_{r}}*a_{c_{r}}}{a_{c_{l}}+a_{c_{r}}}\quad a = \frac{a_{c_{l}}+a_{c_{r}}}{2}
\label{eq:aggregation}
\end{equation}
\vspace{.1cm}

$p_{c_{l}}$ and $p_{c_{r}}$ represent the sentiment polarity labels of the left and right sub-tree respectively. $a_{c_{l}}$ and $a_{c_{r}}$ represent the importance scores, retrieved from the internal MIL attentions. $p$ and $a$ are the respective labels for the parent sentiment polarity and importance score \cite{huber2019predicting}.

As extensively described in \newcite{huber2020MEGA}, the unconstrained CKY approach is not directly applicable for long documents (considered especially important in this work), since the spatial complexity of the CKY approach grows according to the Catalan number, with respect to the number of EDUs in a document. This effectively renders the unconstrained CKY approach insufficient for processing documents with over $\approx 20$ EDUs, even on modern infrastructures\footnote{We used an Intel Core i9-9820X (10 Cores, 3.30 GHz) with a RTX 2080 Ti (128 GB RAM) for our experiment.}. To overcome this problem, we apply the augmentations proposed in \newcite{huber2020MEGA}, reducing the spatial complexity through the application of a beam-search approach, improving the diversity in low-level trees through a stochastic extension. Further, we compute the additional nuclearity attribute, which has previously shown to be an important cue for a variety of downstream tasks \cite{marcu2000theory,ji2017neural,shiv2019novel}. With these extensions, the discourse-tree generation process can be effectively applied to documents of arbitrary length.

\subsection{From Discourse to Sentiment
}
\label{dis_to_sent}


Discourse structure can be 
beneficial and complementary to sequential information for sentiment prediction, especially for long, complicated and nuanced documents (see Figure \ref{fig:example}).
We therefore take a balanced approach in this work, combining a sequential and tree-structured component to predict sentiment. 
Following the intuition by \newcite{bowman2016fast} and \newcite{choi2018learning}, we encode low-level representations in a sequential manner and use the inferred trees on higher levels to guide the prediction of the document-level sentiment.

\begin{figure}[t!]
    \centering
    \includegraphics[width=.7\linewidth]{ 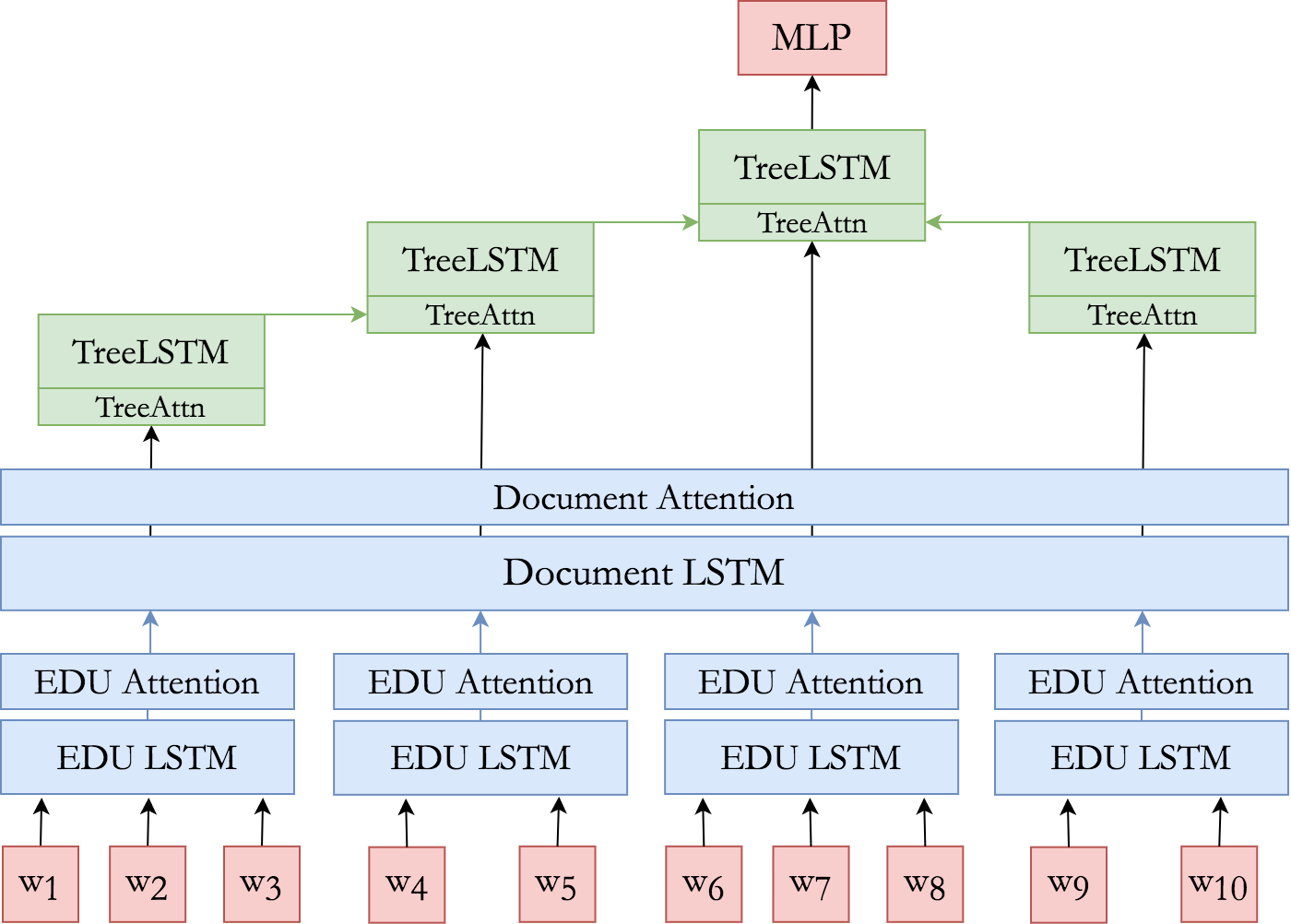}
    \caption{Topology of our hybrid approach using sequential HAN components (blue) in combination with an attention-extended discourse-inspired TreeLSTM (green) aggregation on the dependency discourse tree. Inputs and outputs are red.}
    \label{fig:treelstm}
\end{figure}

\paragraph{Sequential Model Component}
With the HAN model being a strong baseline for many tasks, despite its simple architecture, we decide to take advantage of this contextualization for individual EDUs, as well as for the document-level contextualization (see bottom in Figure \ref{fig:treelstm}). In the standard HAN model the first-level outputs (originally being sentence representations) are used as inputs to a document-level LSTM, augmented with an attention module, to generate the final hidden representation of a document. (see eq. \ref{eq:attn_first} to \ref{eq:attn_third}).\\
\begin{equation}
    u_i = tanh(Wh_i + b)
    \label{eq:attn_first}
\end{equation}
\begin{equation}
    \alpha_i = \frac{exp(u_i^{\top}c)}{\sum_{j \in d} exp(u_j^{\top}c)}
    \label{eq:attn_second}
\end{equation}
\begin{equation}
    h_d = \sum_{i \in d} \alpha_ih_i
    \label{eq:attn_third}
\end{equation}

With $h_i$ as the hidden-state of EDU $i$, obtained from the document-level LSTM, $c$ as the attention context-vector and $d$ representing the set of all sentences/EDUs in the document. We inject discourse information by replacing the computation of the attention weighted sum of the EDU embeddings (equation \ref{eq:attn_third}) with a hierarchical TreeLSTM aggregation of the attention-weighted hidden states. 

\begin{equation}
    h_d = TreeLSTM(\forall_{i \in d} \alpha_ih_i)
    \label{eq:attn_ours}
\end{equation}

We omit the description of the sentence-/EDU-level computations for brevity, as they are unchanged from the original HAN model.

\begin{figure}[t!]
    \centering
    \includegraphics[width=.96\linewidth]{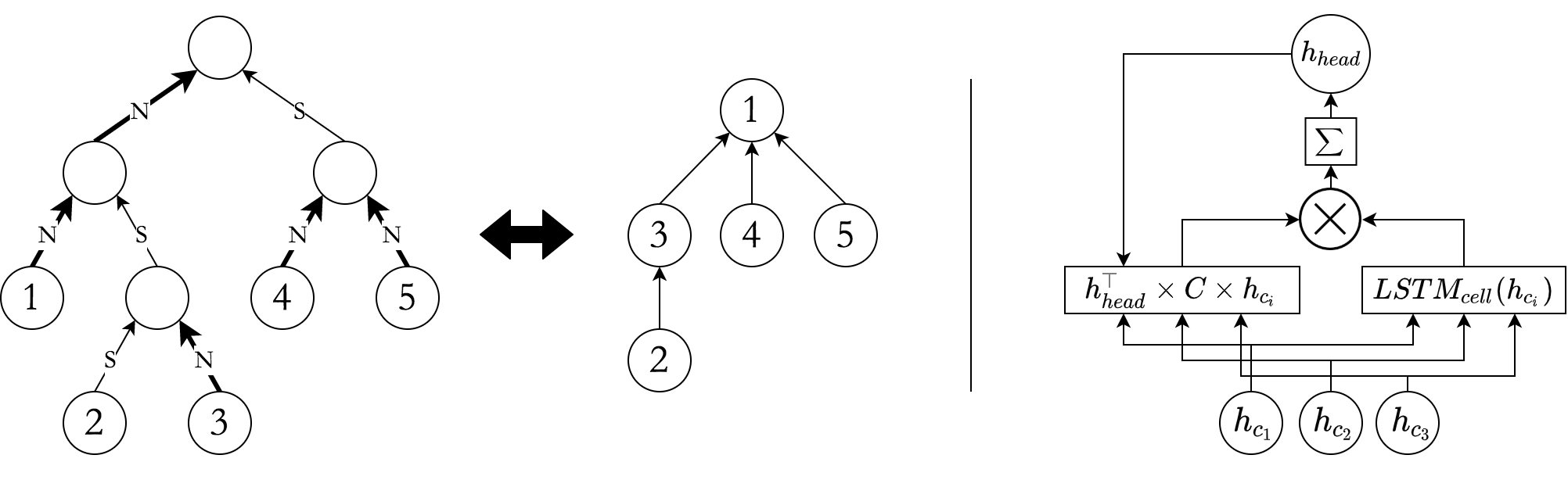}
    \caption{Left: Example transformation of a constituency tree to the respective dependency tree. \\
    Right: Conditional attention module, weighting the importance of child node LSTM encodings $LSTM_{cell}(h_{c_i})$, given the initial head node hidden-state $h_{head}$}
    \label{fig:const_to_det_plus_attention}
\end{figure}

\paragraph{Hierarchical Model Component} 
Using a tree-guided hierarchical aggregation of EDU-level hidden-states to generate a discourse-level hidden representation of the document, we allow more important information according to the discourse tree to be more influential in the computation of the final document representation, as motivated by the examples in Figure \ref{fig:example}. There are two crucial decisions on how to incorporate the discourse-guided tree aggregation: \\
\textbf{(1) The tree representation.} 
Although discourse parsing typically processes constituency tree-structures,
most successful downstream applications of discourse parsing benefit from dependency discourse trees (e.g., \newcite{marcu2000theory}, \newcite{ji2017neural}, \newcite{shiv2019novel}). 
Even though both tree representations are conveying the same information and near-isomorphic conversions are available \cite{morey2018dependency}, we believe that this is because of the different role that nuclearity plays in the tree-representations. 
In particular, while in constituency trees nuclearity is an attribute of internal tree-nodes, head-dependent relations in the dependency tree are fundamentally shaped by the nuclearity attribution. This  more explicit representation of  nuclearity can benefit downstream applications. 
For this reason, we are converting the RST constituency trees into dependency representations (see left of Figure \ref{fig:const_to_det_plus_attention}).

\noindent\textbf{(2) The aggregation approach} has a significant impact on the performance of the model. In this work, we choose the TreeLSTM model by \newcite{tai2015improved}, an evolution of the recursive neural network used in \newcite{ji2017neural}. 
Following the intuition for tree-attention given by \newcite{ji2017neural}, we add a conditional, non-competitive attention module to the child-sum TreeLSTM, augmenting the aggregation of text-spans according to their position in the dependency discourse tree (see eq. \ref{eq:attn_second_tree} to \ref{eq:attn_third_tree}). This extension has not been proposed as part of the TreeLSTM by \newcite{tai2015improved}, however showed improved performance when used in combination with a recursive neural network for the task of discourse parsing \cite{ji2017neural}, which lets us to believe it can also enhance the TreeLSTM for our problem at hand. 

\begin{equation}
    \alpha_i = \sigma(h_{head}^{\top} \times C \times h_{c_i})
    \label{eq:attn_second_tree}
\end{equation}

\begin{equation}
    h_{head} = LSTM_{cell}(\sum_{\mathclap{i \in dep(h_{head})}}\alpha_i h_{c_i})
    \label{eq:attn_third_tree}
\end{equation}

With $C$ as the attention matrix of dimension $(|h_{head}| \times |h_{c_i}|)$, $h_{head}$ representing the hidden-state of the head node and $dep(h_{head})$ 
returning the indices of the dependent child nodes of $h_{head}$.
Please note that the hidden representation of every node in the dependency discourse tree is initialized with the attention-weighted EDU representation obtained from the sequential component and is updated by the TreeLSTM function shown in equation \ref{eq:attn_third_tree}. We combine the head-node EDU representation with the dependants' sub-tree encoding during the bottom-up tree aggregation process (see top of Figure \ref{fig:treelstm} and right of Figure \ref{fig:const_to_det_plus_attention}).
We name our new model DAH (\textbf{D}iscourse \textbf{A}ugmented \textbf{H}AN).

\section{Evaluation}
\label{eval}
In this section, we define the experimental setup and show empirical results of our novel approach, predicting sentiment using sentiment-inspired discourse parsing in the context of previous work. 
We present the datasets used in this work in section \ref{data}. Afterwards, the evaluation metrics and their intuitive justifications are mentioned in section \ref{met}, followed by a short description of the baselines (section \ref{base}). We finish the evaluation section by giving insights into our preliminary evaluations determining the system's hyper-parameters in section \ref{prelim} and describe the final experiments and results in section \ref{final}. 

\subsection{Datasets}
\label{data}
As shown in Figure \ref{fig:stages}, our proposed methodology requires two sets of corpora. In the first step, as described in section \ref{sent_to_dis}, we train a top-performing discourse parser \cite{wang2017two} on a discourse corpus containing RST-style trees. In this step, we use two treebanks:\\
\textbf{RST-DT:} As a human-annotated gold-standard discourse treebank most widely used for discourse related research following the RST theory \cite{mann1988rhetorical}. The dataset contains 385 discourse-annotated news articles from the Wallstreet Journal.\\
\textbf{MEGA-DT:} Our recently proposed ``silver-standard" discourse corpus \cite{huber2020MEGA}, generated in an effort to provide an automatically annotated, large-scale discourse treebank. The corpus is based on the publicly available Yelp'13 sentiment dataset and contains around 250,000 documents annotated with full RST-style discourse trees containing structure and nuclearity attributes. The treebank has shown superior performance to small human-annotated datasets (including RST-DT) on the discourse domain-transfer task, reaching the best performance when evaluated on news/instruction treebanks.

To evaluate the potential of the discourse treebanks to predict sentiment in combination with our novel model architecture, we annotate a large-scale sentiment dataset with discourse trees generated by the discourse parser \cite{wang2017two}, trained on the corpora described above. The publicly available dataset used in this work is the 
\textbf{Yelp'13 dataset}, published by \newcite{tang2015document} 
, containing customer reviews annotated with gold-label sentiment on a 5-point scale. For models incorporating discourse, the previously discourse segmented dataset published by \newcite{angelidis2018multiple} is used with an 80\%/10\%/10\% train/dev/test-split.

Please note that since we use the same base-corpus for training the discourse parser (MEGA-DT) and predicting sentiment for the final evaluation (Yelp'13), we restrict the data used to train the discourse parser to the training-portion of the corpus. This way we ensure that development- and test-documents are unseen during the whole training process.

\subsection{Metrics}
\label{met}
Previous 
models tackle the task of sentiment analysis by interpreting it as a classification problem. While this problem definition is valid for many text categorization tasks, we believe that sentiment analysis 
should be additionally evaluated as a regression task, taking the ordinal nature of the output into account. 
To more 
rigorously evaluate the models in our evaluation, we show four metrics for each system, including the commonly used accuracy and F1-score, as well as the Mean-Squared-Error (MSE) and Mean-Absolute-Error (MAE) metrics. 

\subsection{Baselines}
\label{base}
We compare our new model against two closely related models, namely the Hierarchical Attention Network (HAN) by \newcite{yang2016hierarchical} and the MILNet model \cite{angelidis2018multiple}, which is used as part of the discourse-augmentation process itself in \newcite{huber2019predicting} and \newcite{huber2020MEGA}. With those two closely related baselines we ensure that possible confounding factors in the comparison are minimized, allowing for a clear picture on the effectiveness of incorporating discourse structures into the task of sentiment analysis.

\subsection{Encodings and Hyper-Parameters}
\label{prelim}
To support a fair comparison, we use the same encodings and model-dependent hyper-parameters in all systems.  We replace the domain-depended pre-trained word2vec encodings \cite{mikolov2013distributed} used in the original HAN model, with standard GloVe embeddings \cite{pennington2014glove}. We add MSE and MAE evaluation metrics to the publicly available open-source deep learning toolkit for the original HAN model\footnote{\url{ https://github.com/castorini/hedwig}}. For the MILNet baseline, we align with our previous approach in \newcite{huber2019predicting}, which is also consistent with the adapted HAN model. 
Regarding our novel approach, we convert 
the constituency tree output of the discourse parser into a dependency tree according to \newcite{hayashi2016empirical}.
We run preliminary evaluations on the development-set, comparing a set of loss-function (namely \textbf{Cross-Entropy}, MSE, MAE)\footnote{Selected hyper-parameter is \textbf{bold}} and interpreting the task as either, a classification- or a regression-problem. However, without any further fine-tuning and adaptations, using a regression-based loss is not advisable.
In accordance with the intuition described above, we execute further hyper-parameter search on the main properties of the model itself, exploring a set of 5 learning rates ($\{0.1, 0.05, \textbf{0.01}, 0.05, 0.001\}$) along with three optimization strategies (Adam \cite{kingma2014adam}, AdaGrad \cite{duchi2011adaptive}, \textbf{SGD} \cite{robbins1951stochastic}). We follow the original HAN implementation using $100$ neurons per layer for the bi-directional word and sentence/EDU encodings. The TreeLSTM module contains $512$ neurons. The mini-batch size used in all models is set to $64$, as suggested in \newcite{yang2016hierarchical}. Dropout is set to $50\%$ for all models.

\subsection{Experiments and Results}
\label{final}

\begin{table*}[t!]
\centering
{\renewcommand{\arraystretch}{1.1}
\scalebox{1}{
\begin{tabular}{|l|r r r r|}
\hline
\multirow{2}{*}{Model} & \multicolumn{4}{c|}{Yelp'13}\\
 & Acc & F1 & MSE & MAE\\
\hline \hline 
HAN & 66.20 & 64.26 & 0.486 & 0.379 \\
MILNet & 64.19 & 61.93 & 0.584 & 0.417 \\
DAH\textsubscript{RST-DT} & 65.71 & 63.49 & 0.496 & 0.384 \\
DAH\textsubscript{MEGA-DT} & \textsuperscript{$\simeq\dagger$}66.07 & \textsuperscript{$\simeq\dagger$}64.09 & \textsuperscript{$\simeq\ddagger$}0.491 & \textsuperscript{$\simeq\ddagger$}0.381 \\
\hline
Ensemble(HAN+DAH\textsubscript{MEGA-DT}) & *\textbf{66.27}& *\textbf{64.30} & *\textbf{0.483} & *\textbf{0.377} \\
\hline
\end{tabular}}}
\caption{Final evaluation on the Yelp'13 datasets, subscripts in model names indicate discourse-augmentation treebanks used to generate discourse trees. Best model for each metric is \textbf{bold}. \textsuperscript{$\simeq$}Performance statistically equivalent to HAN model, \textsuperscript{$\dagger$}Discourse-augmentation treebank significantly better than RST-DT with p-value $.05$. \textsuperscript{$\ddagger$}Discourse-augmentation treebank marginally significantly better than RST-DT with p-value $.05$-$.1$, 
*Statistically significant to best model on metric. All significance computations are Bonferroni adjusted.}
\label{tab:final}
\end{table*}

\begin{figure}[t!]
  \centering
  \begin{tikzpicture}
  \begin{axis}[
    width=.7\linewidth, 
    height=6cm, 
    scale only axis,
    anchor=above north west,
    yticklabel pos=right,
    yticklabel=\pgfmathparse{\tick}\pgfmathprintnumber{\pgfmathresult}\,\%,
    ylabel={Accuracy / F1-score},
    xlabel={Document-length range and (support) per Bin},
    xlabel style={at={(0.5,-0.1)}},
    ylabel style={at={(0.07,0.5)}},
    ytick distance=3,
    minor tick num=1,
    xtick=data,
    xticklabels={
        1-211 (27447),
        212-422 (4916),
        423-631 (869),
        633-838 (187),
        848-1054 (85)
        },
    xticklabel style={text width=45},
    xticklabel style={align=center},
    legend pos=south west,
    ymajorgrids=true,
    xmajorgrids=true,
]

    

    \addplot[blue,mark=*]
        coordinates {
            (0, 66.09)	(1, 65.867)	(2, 66.192)	(3, 66.906)	(4, 64.26)

        	};
    \addlegendentry{Acc MEGA-DT}  
    
    \addplot[blue,mark=square*]
        coordinates {
        	(0, 63.618)	(1, 65.133)	(2, 65.488)	(3, 64.987)	(4, 60.478)
        	};
    \addlegendentry{F1 MEGA-DT}
    
    \addplot[brown,mark=*]
        coordinates {
    	    (0, 65.768)	(1, 66.392)	(2, 66.825)	(3, 65.85)	(4, 62.943)
        	};
    \addlegendentry{Acc RST-DT}
    
    \addplot[brown,mark=square*]
        coordinates {
    	    (0, 63.261)	(1, 65.902)	(2, 66.107)	(3, 64.524)	(4, 57.309)
        	};
    \addlegendentry{F1 RST-DT}

    \addplot[red,mark=*]
        coordinates {
        	(0, 66.291)	(1, 66.793)	(2, 66.593)	(3, 65.917)	(4, 61.957)
        	};
    \addlegendentry{Acc HAN}
    
    \addplot[red,mark=square*]
        coordinates {
        	(0, 64.039)	(1, 66.104)	(2, 65.993)	(3, 65.573)	(4, 55.514)
        	};
    \addlegendentry{F1 HAN}
    
    \end{axis}    
\end{tikzpicture}
\caption{Accuracy and F1-score over document-lengths aggregated into 5 bins on the Yelp'13 dataset}
\label{fig:yelp_results_class}
\end{figure}
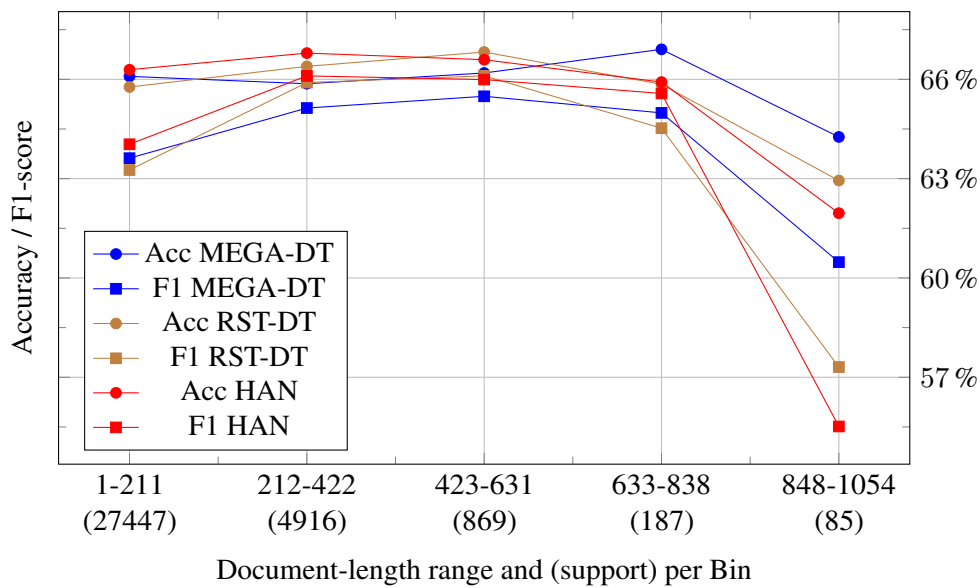

We 
compare 
our novel model using multiple discourse representations obtained from sentiment-inspired discourse structures and standard treebanks against discourse-agnostic systems, solely based on sequential representations on word- and sentence-level. As motivated in Figure \ref{fig:example}, we believe that discourse information is especially useful for long documents, where sentiment is generally expressed in a more diverse or subtle way as compared to short reviews with mostly a clear positive or negative sentiment. We align our evaluation with this intuition by comparing the systems' overall performance in Table \ref{tab:final} and further showing insights into the performance based on the document length in Figure \ref{fig:yelp_results_class}.

The final comparison in Table \ref{tab:final} reports the performance of two baseline systems, not taking discourse information into account, along with two versions of our novel approach, incorporating discourse, and an ensemble method. The performance of all models is averaged over 5 independent runs with different random initializations. All models using discourse (DAH\textsubscript{RST-DT}, DAH\textsubscript{MEGA-DT} and the ensemble of HAN and DAH\textsubscript{MEGA-DT}) are trained with the top-performing discourse parser by \newcite{wang2017two}. All discourse-inspired models further employ an identical neural network architecture, allowing us to directly evaluate the impact of different types of discourse trees on the task of sentiment analysis. 

The best average performance of an individual model (not using the ensemble method) 
is achieved by the sequential HAN model shown in the first row in Table \ref{tab:final}. Even though the average result over 5 independent runs for the DAH\textsubscript{MEGA-DT} system is below the HAN performance, they are statistically equivalent. When compared to the discourse-inspired DAH\textsubscript{RST-DT} model, the performance increase of DAH\textsubscript{MEGA-DT} is statistically significant on the accuracy and F1-score measures and marginally significant for the MSE and MAE. Interestingly, the MILNet model, which is used as an early part of the pipeline to generate the MEGA-DT discourse treebank, does perform substantially worse than the DAH\textsubscript{MEGA-DT} model, which leads us to believe that the combination of the CKY tree aggregation and the DAH sentiment neural-network are able to extract valid and important sentiment information and improve the performance despite the potential propagation of error from the early stage MILNet component. 
Besides the individual models, we also employ an additional experiment with a model-ensemble combining the two top performing models (HAN and DAH\textsubscript{MEGA-DT}), taking their respective strength in different document-length-ranges (as revealed in Figure \ref{fig:yelp_results_class}) into account. The model will be explained in more detail below.

The results shown in Table \ref{tab:final} indicate equal performance of our new DAH\textsubscript{MEGA-DT} methodology when compared to the original HAN model.
However, discourse should arguably be more useful for long documents. Therefore, we further investigate into the document-length dependent performance of the models 
by splitting the test-set into 5 test-document length-depended bins to show the performance across different document sizes (measures by the number of words). We exclude the MILNet baseline in this evaluation due to its clearly inferior performance compared with the sequential HAN model as shown in Table \ref{tab:final}.

The results shown in Figure \ref{fig:yelp_results_class} confirm our initial intuition on the usefulness of discourse structures for long documents, showing strong improvements for our discourse-dependent system in the two rightmost bins. While the performance generally drops for longer documents, the performance decrease is more severe for the sequential HAN model. Generally, we believe that the task of sentiment prediction is harder on longer and more diverse documents, however, we also partly account the performance decrease to the small number of long documents in the Yelp'13 corpus, as shown in the support for each of the bins on the horizontal axis of Figure \ref{fig:yelp_results_class}. While the support shown here is on the test-portion, the general length-distribution on the training- and development-set are similarly skewed towards short documents. 

It can further be seen that the significant performance increase on the overall dataset achieved by the DAH\textsubscript{MEGA-DT} over the DAH\textsubscript{RST-DT} can be mostly attributed to the performance increase in the two right-most bins, containing documents with more than 632 words.

With this confirmation of our initial intuition, we generate a document-length-dependent ensemble of the two top-performing models (HAN and DAH\textsubscript{MEGA-DT}) as mentioned above, to take advantage of the  strength of both systems by selecting the appropriate classifier with a simple threshold -- the document length. To determine the threshold, we evaluate both models on the development-set and select the average of the optimal threshold over 3 runs independently for each metric of interest. 
We then combine the results of the two top performing models on the test-set according to the determined threshold. 
As shown in Table \ref{tab:final}, our ensemble approach significantly outperforms all the individual models, but admittedly only by a narrow margin. Nevertheless, overall the results indicate potential for further improvements in discourse-inspired sentiment analysis for long documents as well as in using ensembles of sequential and tree-driven models to effectively process documents with different levels of complexity.




\section{Conclusion and Future work}
In this work, we explore the next step along the recent line of research on discourse-inspired sentiment analysis, going from sentiment annotations to sentiment prediction through discourse augmentation. We integrate modern discourse parsing approaches into existing, sequential sentiment analysis frameworks, enhancing the model performance through the use of the large-scale MEGA-DT 
discourse dataset and a hybrid approach based on sequential and tree-based components (HAN combined with TreeLSTM). Our proposed approach shows to be especially beneficial when predicting sentiment for long documents containing mixed aspects, combined with complex rhetorical structures. Generating a model-ensemble 
with a simple threshold, based on the document length, improves the overall performance, showing statistically significant results.

We compare our newly developed model with the well-established HAN model. In future work, we plan to compare the standard DocBERT model \cite{adhikari2019docbert} and discourse-inspired versions of it, to further solidify the findings in this work. We also plan to generate other large-scale datasets according to \newcite{huber2020MEGA} and evaluate our model on further ``silver-standard" discourse treebanks. Using a neural discourse parser, such as \newcite{yu2018transition} or \newcite{guz2020roberta} to train on MEGA-DT is another extension of this work.
Besides the task of sentiment analysis, extractive summarization has recently been shown to align well with discourse structures in a transformer framework \cite{xiao2020exsumdiscourse}, giving rise to potential improvements using the DAH model on this task.
As another extension, we intend to look into more sophisticated ways 
to ensemble the sequential- and discourse tree-based models.
\section*{Acknowledgments}
We thank the anonymous reviewers and the UBC-NLP group for their insightful comments and suggestions. 
This research was supported by the Language \& Speech Innovation Lab of Cloud BU, Huawei Technologies Co., Ltd.

\bibliographystyle{coling}
\bibliography{coling2020}

\clearpage

\appendix
\onecolumn
\section{Sentiment Annotated Discourse Trees for long Documents}
\label{appendix_1}
\begin{figure}[ht!]
    \centering
    \includegraphics[width=1\linewidth]{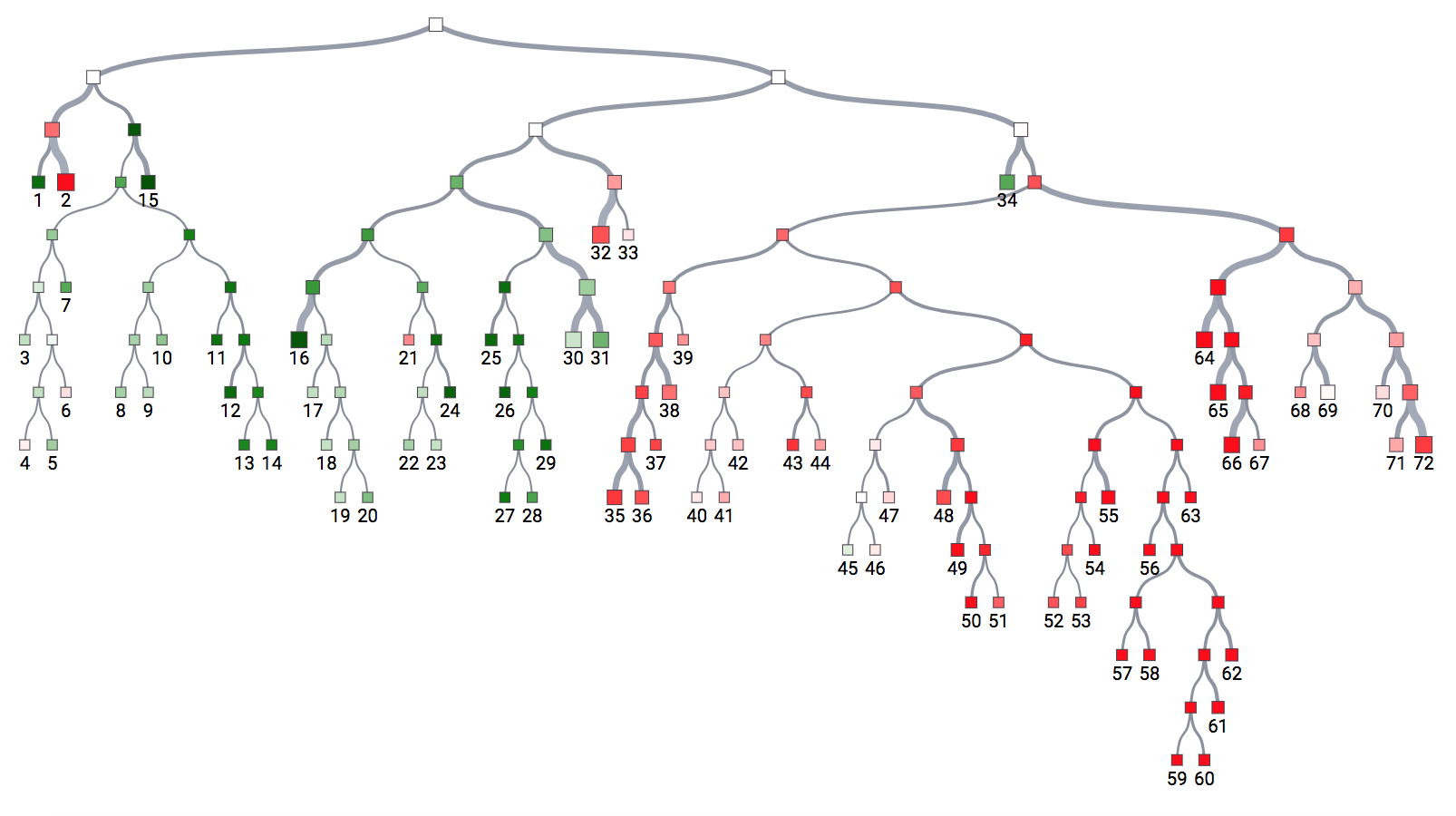}
    \caption{Discourse: [amazing food.]$_{1}$, [awful, awful service.]$_{2}$, [the garlic bread. very good.]$_{3}$, [softer than i expected,]$_{4}$, [which was nice.]$_{5}$, [i also just wasn't expecting garlic bread.]$_{6}$, [so it was a nice surprise.]$_{7}$, [escargot -]$_{8}$, [i was the only one at the table (of 10)]$_{9}$, [to eat it.]$_{10}$, [they were great!]$_{11}$, [served bubbling hot, not rubbery at all, delicious sauce.]$_{12}$, [i kept the dish]$_{13}$, [to dip bread into just because of the sauce.]$_{14}$, [veal - amazing.]$_{15}$, [everything tasted fantastic.]$_{16}$, [ok, the carrots]$_{17}$, [that were on the side were a bit plain]$_{18}$, [and could have been softer, but the veal itself and the sauce]$_{19}$, [it was in, and the mushrooms and pasta.]$_{20}$, [i left nothing on my plate.]$_{21}$, [my husband got the same]$_{22}$, [and also had the same impression.]$_{23}$, [creme brulee - fantastic.]$_{24}$, [tasted great, good texture.]$_{25}$, [pleasantly surprised.]$_{26}$, [my husband got the tiramisu]$_{27}$, [and said]$_{28}$, [it was great.]$_{29}$, [so why the 3 stars]$_{30}$, [when the food was so amazing?]$_{31}$, [because of the terrible service. 1 -]$_{32}$, [we got water.]$_{33}$, [great.]$_{34}$, [but our server * never * asked us]$_{35}$, [if we wanted anything else.]$_{36}$, [when my husband finally stopped him to ask for a glass for my father in law, a coke for]$_{37}$, [and other drinks, our server looked very inconvenienced by it. 2 -]$_{38}$, [didn't get to order appetizers.]$_{39}$, [you see]$_{40}$, [i got escargot?]$_{41}$, [i ordered that with my meal.]$_{42}$, [our server never asked about appetizers]$_{43}$, [and went straight to meals.]$_{44}$, [also, my husband was walking with our daughter]$_{45}$, [when the ordering was starting]$_{46}$, [and needed an extra minute.]$_{47}$, [our server wanted to start with him.]$_{48}$, [when asked if he could start with someone else's order,]$_{49}$, [our server protested,]$_{50}$, [but eventually did move on to the next person.]$_{51}$, [you'd think]$_{52}$, [starting at the next person was]$_{53}$, [asking him to cut off his hand. 3 - empty glasses everywhere!]$_{54}$, [never got or was offered a refill on my drink.]$_{55}$, [or anyone else's.]$_{56}$, [when my father stopped our server well]$_{57}$, [after our meal was over]$_{58}$, [and asked]$_{59}$, [if i could get a coke,]$_{60}$, [our server said]$_{61}$, [i had never ordered one.]$_{62}$, [well of course i hadn't.]$_{63}$, [i never had a chance to! 4 -]$_{64}$, [offering dessert seemed a complete afterthought.]$_{65}$, [will i recommend this place to anyone else?]$_{66}$, [conditionally.]$_{67}$, [i'll make sure to tell them]$_{68}$, [that the food was very good, but not to go]$_{69}$, [if they want attentive service,]$_{70}$, [are on any kind of time constraint, expect refills on their drinks,]$_{71}$, [or are at all shy about getting a server's attention.]$_{72}$}
\end{figure}

\end{document}